\documentclass{article}

\usepackage{microtype}
\usepackage{graphicx}
\usepackage{subfigure}
\usepackage{booktabs} 
\usepackage{eucal}
\usepackage{bbm}
\usepackage{hyperref}

\usepackage{amsmath,amsfonts,bm}

\def\eqref#1{equation~\ref{#1}}

\def\1{\bm{1}}

\DeclareMathAlphabet{\mathsfit}{\encodingdefault}{\sfdefault}{m}{sl}
\SetMathAlphabet{\mathsfit}{bold}{\encodingdefault}{\sfdefault}{bx}{n}

\usepackage[accepted]{icml2024}

\usepackage{amsmath}
\usepackage{amssymb}
\usepackage{mathtools}
\usepackage{amsthm}
\usepackage{enumitem}
\usepackage{subfigure}
\usepackage{caption}
\usepackage{subcaption}

\usepackage[capitalize,noabbrev]{cleveref}
\usepackage{xspace}

\theoremstyle{plain}

\theoremstyle{definition}

\theoremstyle{remark}

\icmltitlerunning{Decomposed evaluations of geographic disparities in text-to-image models}

\definecolor{darkgreen}{rgb}{0.0, 0.5, 0.0}
\definecolor{royalblue}{RGB}{50, 92, 168}

\newcommand{\methodname}{Decomposed-DIG\xspace}
\newcommand{\methodnamelong}{Decomposed Indicators of Disparities in Image Generation\xspace}
\newcommand{\bgonly}{BG-only\xspace}
\newcommand{\objonly}{Obj-only\xspace}
\newcommand{\fullimg}{Full-image\xspace}

\begin{document}

\twocolumn[
\icmltitle{Decomposed evaluations of geographic disparities in text-to-image models}



\icmlsetsymbol{equal}{*}
\icmlsetsymbol{seniorequal}{**}

\begin{icmlauthorlist}
\icmlauthor{Abhishek Sureddy}{equal,yyy}
\icmlauthor{Dishant Padalia}{equal,yyy}
\icmlauthor{Nandhinee Periyakaruppa}{equal,yyy}
\icmlauthor{Oindrila Saha}{yyy}
\icmlauthor{Adina Williams}{Meta}
\icmlauthor{Adriana Romero-Soriano}{Meta,Mila,McGill,CIFAR} 
\icmlauthor{Megan Richards}{seniorequal,Meta}
\icmlauthor{Polina Kirichenko}{seniorequal,Meta,NYU}
\icmlauthor{Melissa Hall}{seniorequal,Meta}

\end{icmlauthorlist}

\icmlaffiliation{yyy}{Department of Computer Science, University of Massachusetts Amherst}
\icmlaffiliation{Meta}{FAIR Labs, Meta AI}
\icmlaffiliation{NYU}{New York University}
\icmlaffiliation{Mila}{Mila-Quebec AI Institute}
\icmlaffiliation{McGill}{McGill University}
\icmlaffiliation{CIFAR}{Canada CIFAR AI Chair}

\icmlcorrespondingauthor{Melissa Hall}{melissahall@meta.com}

\vskip 0.3in
]

\setlength{\parskip}{0.1cm plus1mm minus1mm}

\printAffiliationsAndNotice{\icmlEqualContribution \icmlEqualSeniorContribution} 

\begin{abstract} 
\vspace{0.2mm}
    Recent work has identified substantial disparities in generated images of   different geographic regions, including stereotypical depictions of everyday objects like houses and cars. 
    However, existing measures for these disparities have been limited to either human evaluations, which are time-consuming and costly, or automatic metrics evaluating \textit{full} images, which are unable to attribute these disparities to specific parts of the generated images. 
    In this work, we introduce a new set of metrics, \methodnamelong (\methodname), that allows us to separately measure geographic disparities in the depiction of \textit{objects} and \textit{backgrounds} in generated images. 
    Using \methodname, we audit a widely used latent diffusion model and find that generated images depict objects with better realism than backgrounds and that backgrounds in generated images tend to contain larger regional disparities than objects. 
    We use \methodname to pinpoint specific
    examples of disparities, such as stereotypical background generation in Africa,
    struggling to generate modern vehicles in Africa,
    and unrealistically placing some objects in outdoor settings.
    Informed by our metric, we use a new prompting structure that enables a 52\% worst-region improvement and a 20\% average improvement in generated background diversity.\footnote{All experiments were conducted by students at the University of Massachusetts Amherst on University of Massachusetts Amherst servers.}
    
\end{abstract}

\section{Introduction}

\begin{figure}[ht]
    \centering
    \begin{minipage}[t]{0.49\textwidth}
        \centering
        \includegraphics[width=
    \textwidth]{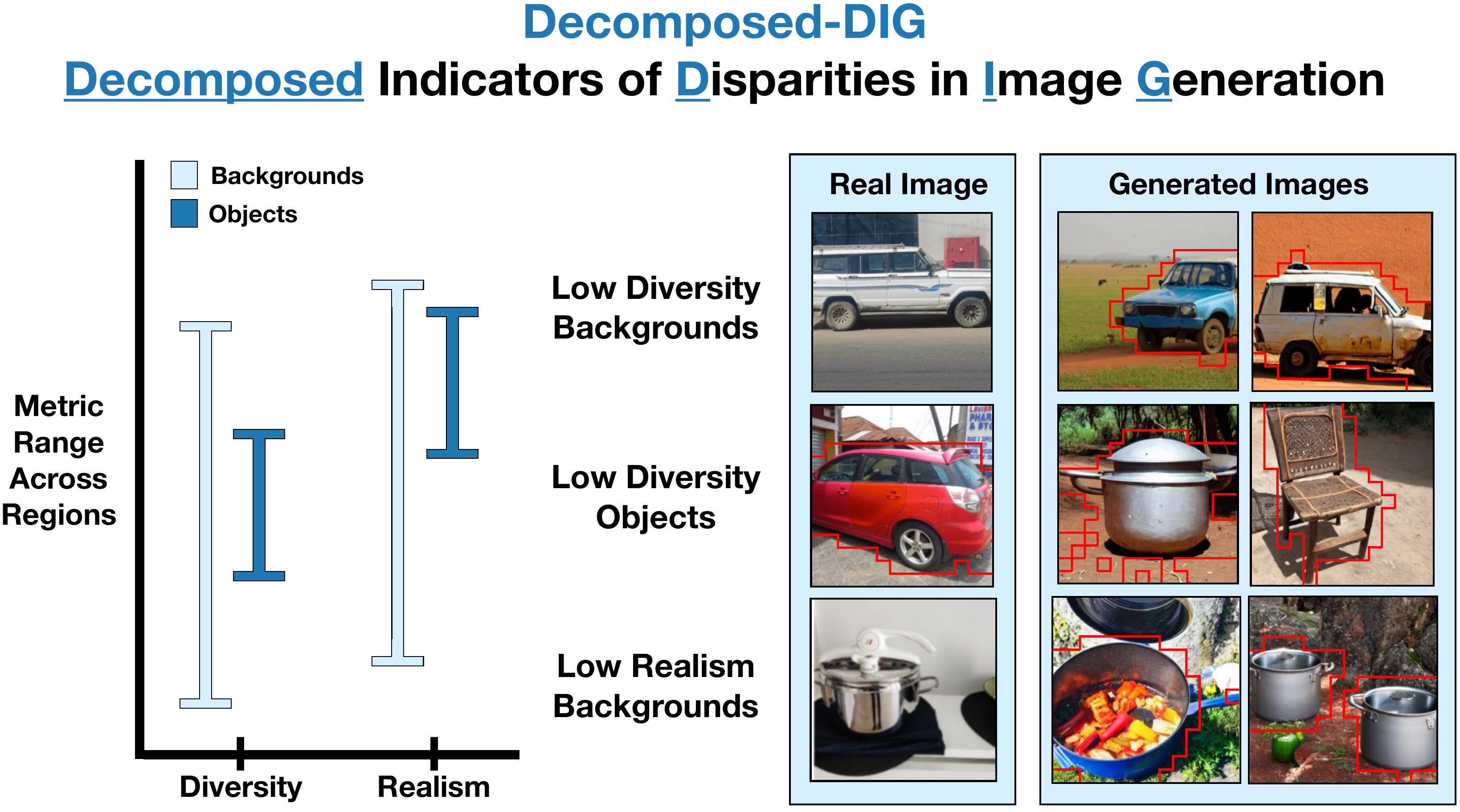}
    \end{minipage}
    \caption{We introduce \methodname, which decomposes measurements of geographic disparities in text-to-image generation between \textit{object} and \textit{background} representations. Using \methodname, we identify generation patterns that contribute to geographic disparities.}
    \label{fig:SummaryFig1}
\end{figure}

Recent advancements in text-to-image generative systems have driven immense progress both for visual content creation \cite{Rombach_2022_CVPR,ramesh2022hierarchical} and training downstream discriminative models \cite{hemmat2023feedbackguided,tian2023stablerep,yeo2024controlled}. 
Despite this progress, there has also been increasing evidence that they produce content containing social biases that do not depict the real world accurately. 

For example, recent works have identified that generated images often contain geographic biases, with large disparities in image realism and representation diversity among geographic regions \cite{hall2023dig}. 
These models can amplify harmful stereotypes, such as producing images that over-represent poverty and rudimentary infrastructure for Africa \cite{bansal2022how} or failing to produce adequate geographic representation  for some regions \cite{basu2023inspecting}. 
To evaluate these disparities, some works propose automatic metrics that evaluate generated images \textit{holistically}, either in comparison to real world images \cite{hall2023dig} or via group association rates \cite{lee2023holistic}.
However, recent work shows that humans interpret geographic representation via specific \textit{components} of images, such as buildings presented in the background, natural vegetation, or stylization of central objects \cite{hall2024geographic}. Therefore, it is important to understand which segments of generated images contribute to disparities to better inform the development of mitigations. 

As a step in this direction, we introduce a decomposed evaluation protocol to disentangle and measure disparities between the target concept and its accompanying background in generated images. 
We (1) extend the precision- and coverage-based Disparity in Image Generation (DIG) indicators proposed in \citet{hall2023dig} and decompose them into object- and background-indicators and (2) enhance them by leveraging a state-of-the-art segmentation method \citep{kirillov-etal-2023-segment}.
We call this set of metrics ``\methodname.'' 

Using \methodname, we uncover a much more nuanced picture of geographic bias in model-generated images. 
In particular, we study images generated with the prompt \texttt{\{object\} in \{region\}} and find that:
\begin{itemize}[noitemsep,topsep=0pt,parsep=0pt,partopsep=0pt]
    \item Generated images tend to exhibit higher realism for objects than backgrounds.
    \item The depiction of backgrounds in generated images has $1.7$x larger disparities between geographic regions than the depiction of objects.
    \item \methodname enables more precise characterization of bias modes of generative models, such as rarely including paved streets or buildings in backgrounds of images of Africa. 
\end{itemize}

Informed by our findings, we explore prompting as an early mitigation, finding that a prompt template that defines geographic information as an adjective, i.e.\ ``European car,'' improves background diversity by 52\% for the worst-performing region and by 20\% on average, with slight improvement or little cost to background realism and object representation. We hope this work encourages more fine-grained, realistic evaluation of generative vision models and informs mitigations to better enable accurate, representative generations for all global regions. 

\begin{figure*}[ht]
    \centering
    \begin{minipage}[t]{0.51\textwidth}
        \centering
        \vspace{-0.5\abovecaptionskip}
        \includegraphics[width=\linewidth]{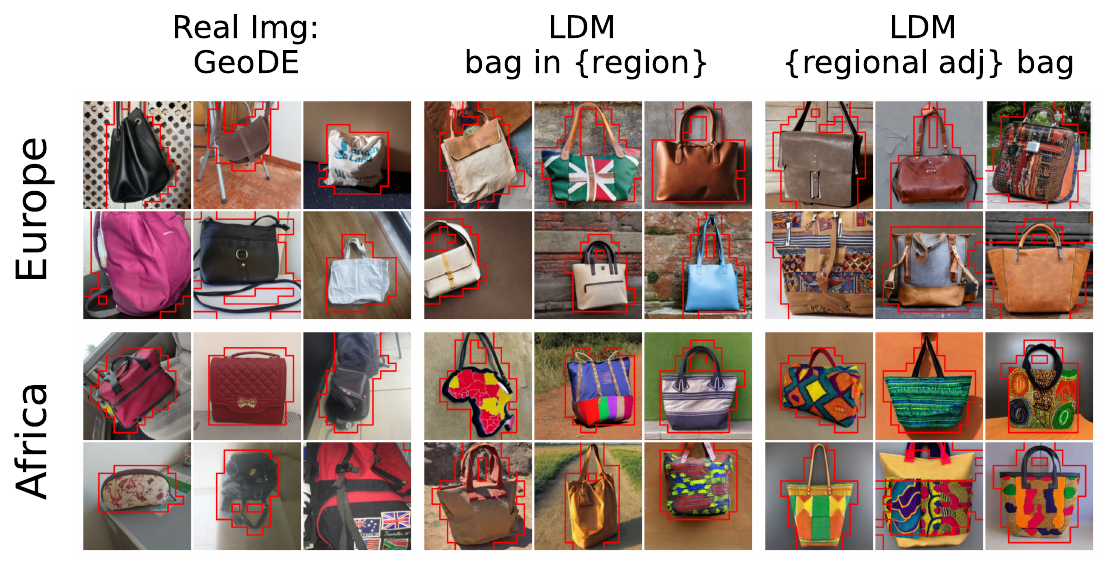}
    \end{minipage}
    \hfill
    \begin{minipage}[t]{0.48\textwidth}
        \centering
        \vspace{-0.5\abovecaptionskip}
        \includegraphics[width=\linewidth]{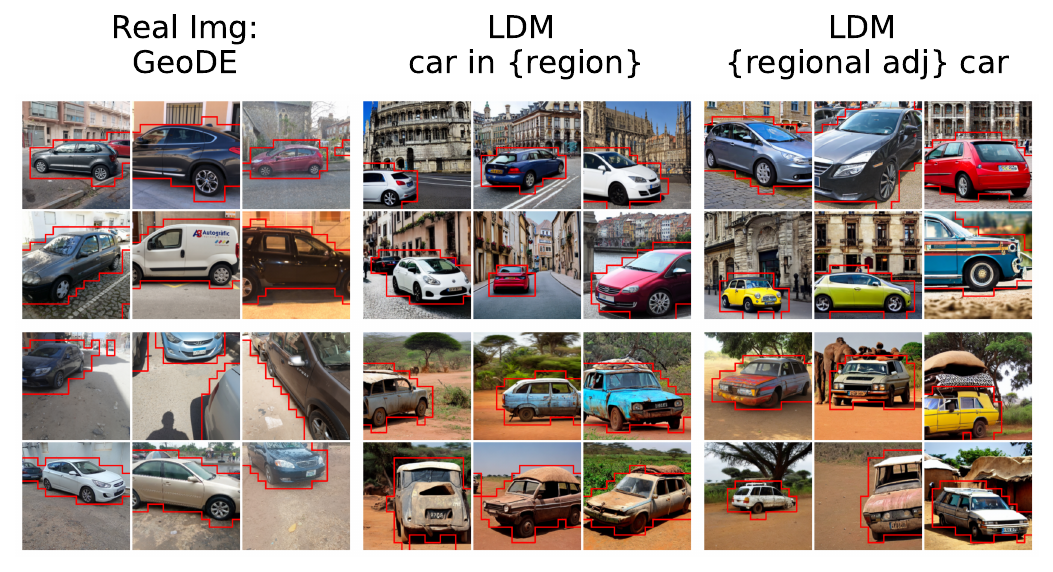}
    \end{minipage}
    \caption{Examples of real and generated images of objects in different regions (Europe in top row and Africa in the bottom row). 
    We propose \methodname to pinpoint geographic disparities related to the depiction of objects and backgrounds in generated images created with the prompt \texttt{\{object\} in \{region\}}. 
    We then study an alternative prompt template that emphasizes the object more than the region: \texttt{\{regional adjective\} \{object\}} i.e. ``European bag'', which leads to higher background diversity.
    Red outlines show object/background decompositions.}
    \label{fig:main-disparities}
\end{figure*}

\section{Background}
\label{sec:background}

We build upon existing indicators for disparities in image generation (DIG) \cite{hall2023dig}.
DIG utilizes a dataset of real images that capture the desired geographic representation as a reference dataset and thousands of images from the text-to-image model being evaluated.
We focus on the Region Indicator, which uses manifold-based metrics precision \cite{kynkäänniemi2019improved} and coverage \cite{naeem2020reliable} to measure geographic disparities in terms of realism and representation diversity, measuring disparities in generations of \textit{objects} in different \textit{geographic regions}.

\subsection{Datasets}
\label{sec:background_images}
\vspace{-0.5mm}
\paragraph{Reference dataset of real images.}
For real images, we use the geographically representative dataset GeoDE \citep{ramaswamy2022geode},  which contains images of $40$ objects in $6$ geographic regions. 
Depictions of objects take up at least $25\%$ of each image, per dataset requirements. Following \citet{hall2023dig}, we balance GeoDE to ensure the same number of images per object-region combination, yielding $27$ objects and $29$k images. 

\vspace{-0.5mm}
\paragraph{Dataset of generated images.}

We study a latent diffusion model (LDM) and use the prompt structure \texttt{\{object\} in \{region\}} as a template. 
We include all object-region combinations to match the distribution of images present in our filtered version of GeoDE. 

\subsection{DIG Indicators}

The existing DIG Indicators \cite{hall2023dig} measure disparities in the realism and diversity of the generated images across different regions by utilizing precision~\cite{kynkäänniemi2019improved} and coverage scores~\cite{naeem2020reliable}.
We summarize these metrics below and include their full definition in Appendix \ref{sec:app_methodology}.

\vspace{-0.5mm}
\paragraph{Precision.}
Precision \cite{kynkäänniemi2019improved} approximates the \textit{realism} of generated images by measuring the proportion of generated images that fall close to the set of real images. 
In particular, both the real and generated images are mapped to a shared feature space.
A manifold of real images is constructed from the hypersphere of each real image, where the hypersphere is defined by the distance to the third-nearest neighbor of the respective image. 
Precision thus measures the proportion of generated image features that fall into the real image manifold. 
Intuitively, a higher precision value means that there are more generated images that are visually similar to a real image, i.e.\ more realistic.

\vspace{-0.5mm}
\paragraph{Coverage.}
Coverage \cite{naeem2020reliable} approximates the \textit{diversity} of generated images. 
We use the same image features described above and count the proportion of real images for which at least one generated image falls in its hypersphere. 
Thus, a higher coverage indicates that more real images have representation among the generated images. 

\section{\methodname}

We now introduce our benchmarking protocol \methodnamelong (``\methodname'').
Assuming one has access to real and generated images (see  \Cref{sec:background_images}), \methodname requires three steps: (1) segment objects and backgrounds from images, (2) extract decomposed features, and (3) perform object- and background-specific measurements. 

\subsection{Object and background segmentation}

To divide each image into object and background components, we perform segmentation using the Segment Anything Model (SAM; \cite{kirillov-etal-2023-segment}).
However, because SAM requires points, bounding boxes, or pre-existing masks as input prompts, we use the LangSAM library\footnote{https://github.com/luca-medeiros/lang-segment-anything} which leverages GroundingDINO's \cite{liu2023grounding}  zero-shot bounding box object detection as inputs to SAM.
We use bounding boxes corresponding to the specific object that we know appears in the real image or was used in prompting for the generated image.
SAM then yields precise segmentation
masks for object class. 
We consider the remaining image regions as image backgrounds.

In our experiments, we find that LangSAM works well for nearly all object classes in real images, exhibiting failures for a small subset of object classes such as ``cleaning equipment,'' ``hand soap,'' and ``light fixture'', which we remove from our evaluation (see Appendix \ref{sec:app_methodology} for more details).\looseness-1

\subsection{Decomposed image features}

With the object and background segmentation, we use a vision transformer (ViT) \citep{dosovitskiy2020image} for feature extraction
(specifically, ViT base model with $16$x$16$ patching pre-trained on ImageNet-21K and fine-tuned on ImageNet-1K). 
Unlike
CNN-based feature extractors such as InceptionV3 \cite{DBLP:journals/corr/SzegedyVISW15},
the ViT's use of patches allows us to isolate the features corresponding to objects and backgrounds \citep{jain2022missingness}. Using the segmentation mask from the previous step, we consider all the patches that contain at least one pixel of the object as object patches, and the remaining patches (which do not have any object pixels) as background patches (thus, together they make up the full image without overlap).
For object-specific measurements, we mask out background patches by zeroing out the attention scores corresponding to these patches during the forward pass.
Similarly, for background-specific measurements, we zero out the attention scores of the object patches. For full-image measurements, we use all the patches without masking. We use the CLS token features from the last layer of the ViT as the image features for each of the set-ups.

\subsection{Object- and background-specific measurements}

With the object and background ViT features, we benchmark 1) object-specific (``\objonly'') and 1) background-specific (``\bgonly'') disparities in realism and diversity in generated images. 
We also maintain full image measurements (``\fullimg'') from previous work \citep{hall2023dig}, including ViT tokens for the entire image. 
\vspace{-0.5mm}
\paragraph{\objonly benchmarking.}
For object-specific measurements, we disaggregate real and generated images between geographic regions and calculate precision and coverage measurements by using only the ViT features corresponding to \textit{object} segmentations.
The object features coming from real images are used to approximate the manifold of real objects, whereas the object features coming from generated images are projected onto the real object manifold to compute precision and coverage. 
We include generated images that contain no object segmentations in the overall count of samples, which helps in penalizing generations that have consistency issues as observed in previous works \cite{hall2023dig,hall2024geographic}.
\vspace{-0.5mm}
\paragraph{\bgonly benchmarking.}
Similarly, for benchmarking of disparities in depicted backgrounds, we disaggregate between geographic regions and calculate precision and coverage using features corresponding to image \textit{backgrounds}.
In this case, generated images that do not have object segmentations are included in their entirety.

\section{Results}

Next, we leverage Decomposed-DIG to better understand the object and background components of the geographical disparities in the widely used LDM 1.5.\footnote{We focus on \objonly and \bgonly measurements in the main text and include \fullimg results in Appendix \ref{app:additional_findings}.}

\subsection{Objects have better realism than backgrounds}
\label{subsection:obj-more-realism}

\begin{figure}[ht]
    \centering
    \begin{minipage}[t]{0.35\textwidth}
        \centering
        \includegraphics[width=
    0.8\textwidth]{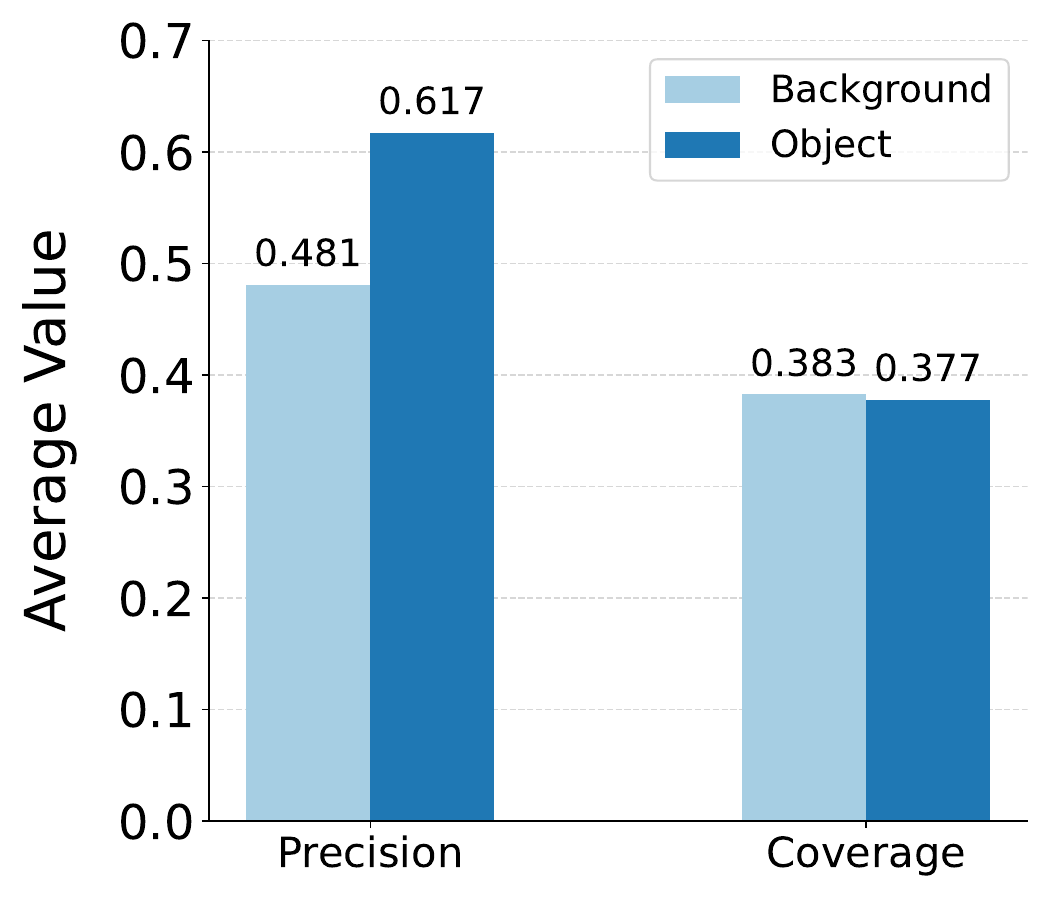}
    \end{minipage}
    
    \caption{\textbf{In generated images, objects tend to have better realism (precision) than backgrounds, while representation diversity (coverage) is similar on average between objects and backgrounds.}
    Shown are precision and coverage measurements averaged over all regions for \objonly and \bgonly set-ups.
    }
    \label{fig:pc_avgd_barplot}
\end{figure}

We first compare the generations of objects and backgrounds, computing \methodname metrics on \objonly and \bgonly portions of generated images, shown in Figure \ref{fig:pc_avgd_barplot}. We find lower precision in \bgonly decompositions compared to \objonly segments, across region subsets. 
This difference in precision suggests that generated backgrounds are less similar to real backgrounds than generated objects are to real objects.
The examples in Figure \ref{fig:main-disparities} are consistent with this finding as generated backgrounds often portray settings that are not frequently found in real images, such as rural dirt scenes for Africa or large stone historic architecture for Europe.
We further investigate the generation patterns that drive this difference in  Section \ref{subsection:bias_modes}.  

In addition, we find that \textit{on average}, coverage is similar between \objonly and \bgonly set-ups. 
However, \textit{per-region} disparities vary notably between the objects and backgrounds, which we discuss next.

\label{subsection:backgrounds-more-disparities}
\begin{figure}[ht]
    \centering
    \begin{minipage}[t]{0.45\textwidth}
        \centering
        \includegraphics[width=
    \textwidth]{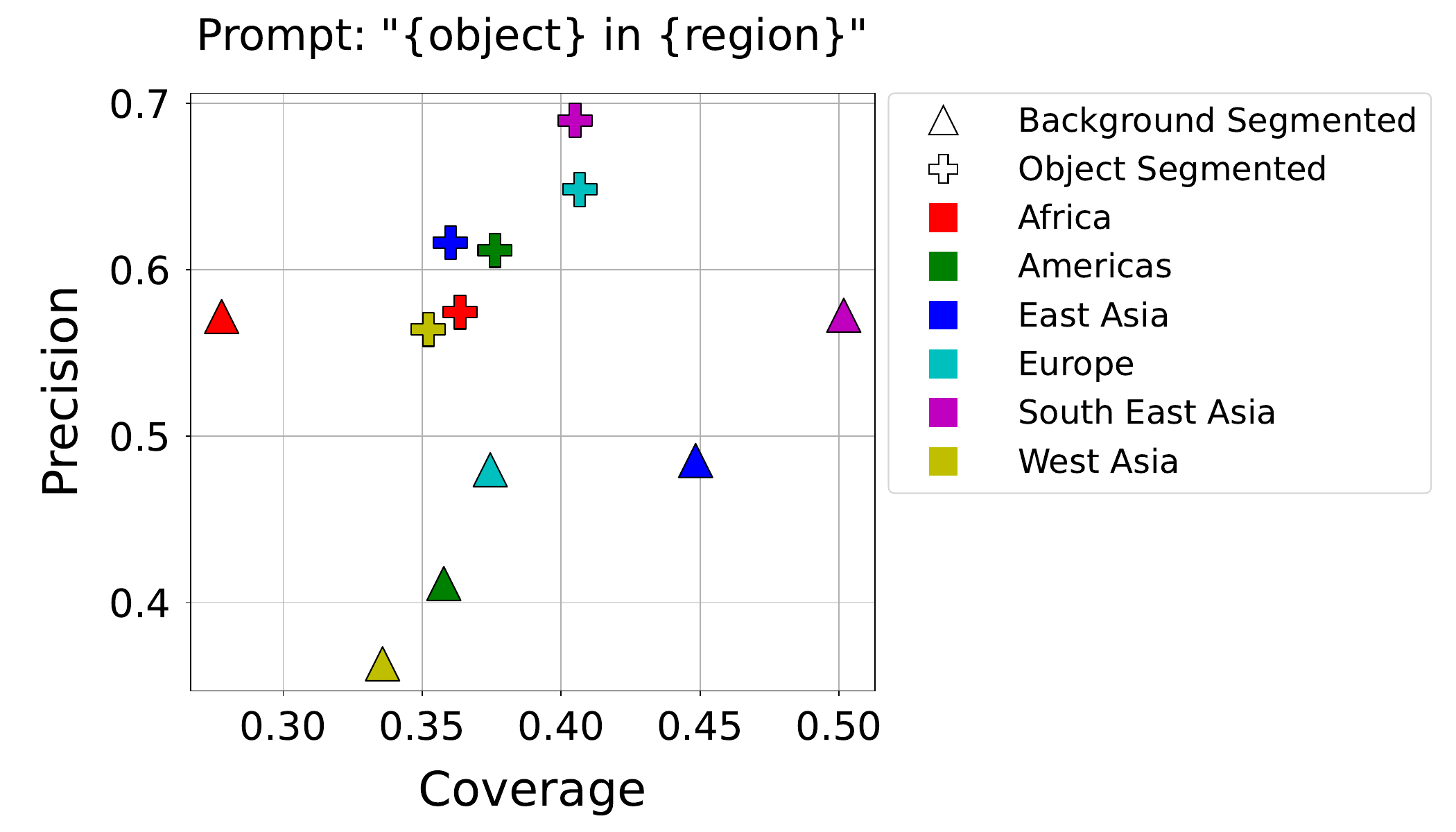}
    \end{minipage}
    \caption{\textbf{Backgrounds in generated images have larger disparities in realism and diversity between geographic regions than objects.}
    We observe the larger variance of precision and coverage values for \bgonly compared to \objonly set-up.
    }
    \label{fig:plot_pc_combined_all_regions_section_4_2}
\end{figure}

\subsection{Backgrounds have larger geographic disparities than objects}
Using \methodname, we find that backgrounds have much larger geographic disparities than objects (shown in Figure \ref{fig:plot_pc_combined_all_regions_section_4_2}).
In particular, \bgonly measurements reveal that background coverage is \textit{twice} as large for the best-performing region (Southeast Asia) than the worst performing region (Africa), while only $1.2$x larger for object coverage. 
Furthermore, we see that per-region precision scores for backgrounds span about $1.5$x the amount of per-region scores for objects, demonstrating larger region disparities for backgrounds.
Finally, we observe that unlike the other regions, Africa shows changes only in coverage (not precision) between object and background setups. 
We discuss this further in Appendix \ref{app:additional_findings}.

\subsection{Characterizing generation patterns contributing to geographic disparities}
\label{subsection:bias_modes}

\begin{figure}[t]
    \centering
    \hfill
    \hfill
    \hfill
    \begin{minipage}[t]{0.228\textwidth}
        \centering
        \includegraphics[width=
    \textwidth]{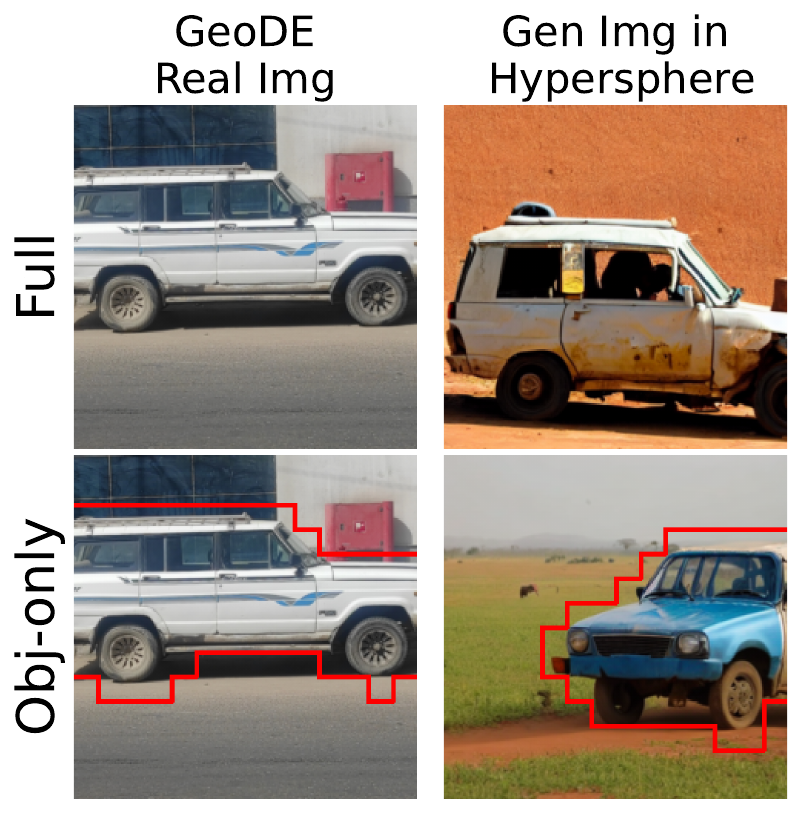}
    \end{minipage}
    \hfill\vline\hfill
    \begin{minipage}[t]{0.21\textwidth}
        \centering
        \includegraphics[width=
    \textwidth]{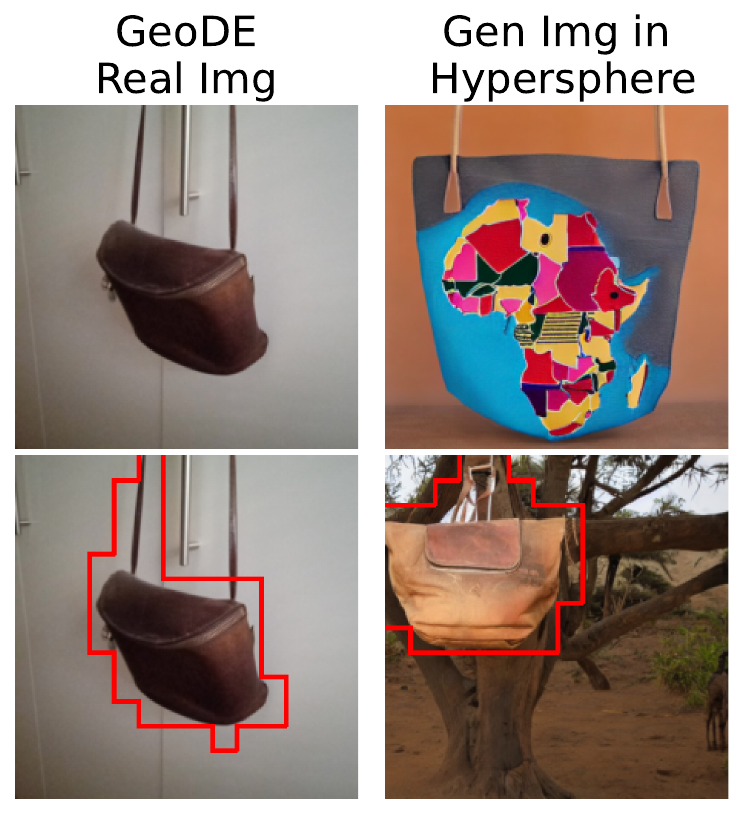}
    \end{minipage}
    \hfill
    \hfill
    \hfill
    \caption{\textbf{When generating objects in Africa, the LDM struggles to depict full background diversity, in particular backgrounds with buildings and paved streets (left) and neutral indoor scenes (right).}
    Depicted are examples of real images where there are no generated images in the hypersphere of nearest neighbors for \bgonly, but there are generated images in the hypersphere for \fullimg and \objonly (shown). 
    Red outlines show object/background decompositions.
    }
    \label{fig:failure1}
\end{figure}

Informed by our quantitative findings, we next use \methodname to qualitatively pinpoint generation patterns that contribute to geographic disparities, investigating low diversity backgrounds, low diversity objects, and low realism backgrounds. 
\vspace{-0.5mm}
\paragraph{Low Diversity Backgrounds: Generated backgrounds of Africa rarely include neutral, grey colored scenes.}
We examine real images pertaining to low diversity background generations, i.e.\ real images contain generated images in their hyperspheres with the \objonly set-up but not with the \bgonly set-up.
This reveals examples where the generated images do not cover the full diversity of real backgrounds.
For example, in Figure \ref{fig:failure1} we show examples of real images of a car and a bag in Africa with a boxy, SUV-like appearance in a neutral background featuring grey concrete and an unmarked building. 
When we consider the full image in application of the original DIG measurements, multiple generated images depicting a similar object but starkly different backgrounds are included in the real image's hypersphere. 
However, when we utilize \methodname and split up the analysis between object and background, we find that there are no generated image backgrounds that fall in the real image's background hypersphere. 
Thus, we identify our first failure mode: the LDM struggles to depict realistic neutral scenes featuring plain grey backgrounds in the generation of objects in Africa.

\vspace{-0.5mm}
\paragraph{Low Diversity Objects: Generated cars in Africa do not include red sedans.}

\begin{figure}[t]
    \centering
    \begin{minipage}[t]{0.199\textwidth}
        \centering
        \includegraphics[width=
    \textwidth]{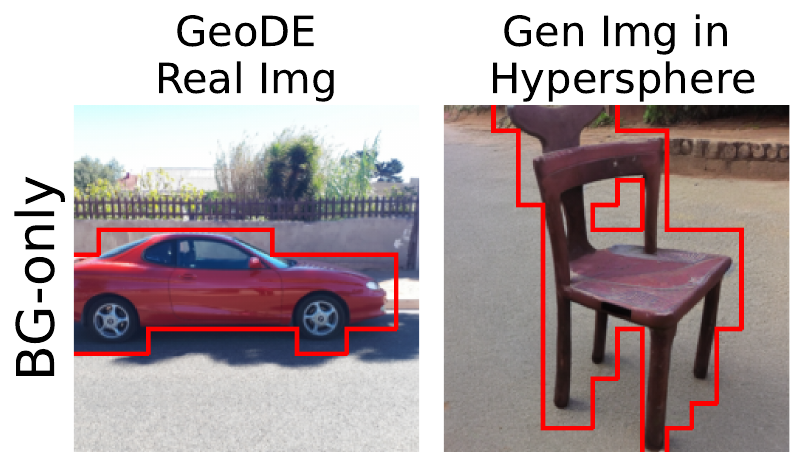}
    \end{minipage}
    \hfill\vline\hfill
    \begin{minipage}[t]{0.275\textwidth}
        \centering
        \includegraphics[width=
    \textwidth]{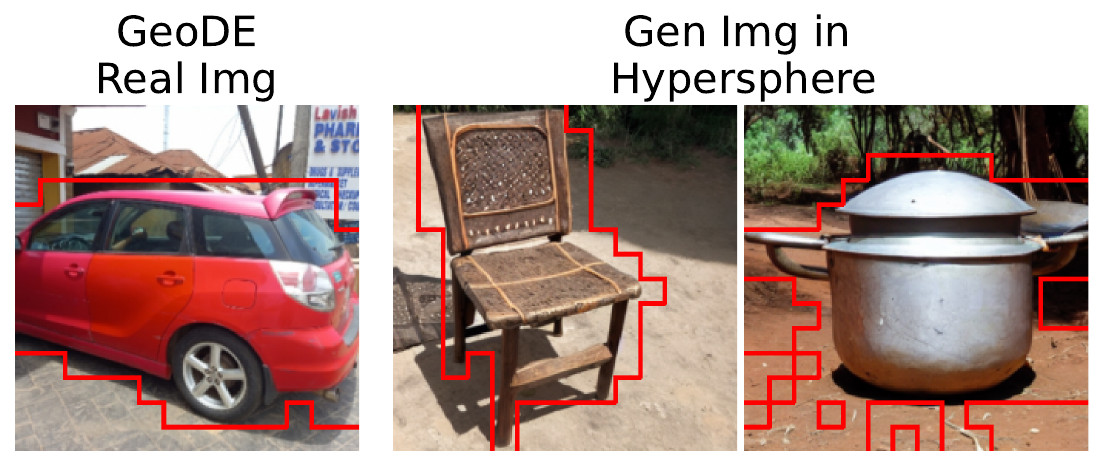}
    \end{minipage}
    \caption{\textbf{The LDM struggles to depict full object diversity, particularly more modern vehicles (such as sedans) in Africa}. Shown are real images with the generated images that are in their hypersphere of nearest neighbors for \bgonly measurement, but not \fullimg or \objonly.
    Red outlines show object/background decompositions.
    ~\looseness-1
    }
    \label{fig:failure2}
    
\end{figure}

We next inspect real images pertaining to low diversity object generations, i.e.\ with generated images in their hyperspheres in the \bgonly set-up but not the \objonly set-up.
These serve as examples where the generations do not cover the full diversity of real objects.
Figure~\ref{fig:failure2} shows examples of real images of red sedans in Africa that have generated images in their hypersphere for the \bgonly set-up but not the \fullimg or \objonly measurement. 
This shows the utility of \methodname in allowing us to pinpoint that the reason for the full image's empty hypersphere is not due to lack of background representation in the generated images but rather object depiction.
And indeed, when we manually inspect all images of generated cars in Africa, we find no examples of red sedans and more than $90\%$ are dirty and rusty (similar to those in Figure \ref{fig:main-disparities}). ~\looseness-1

\vspace{-0.5mm}
\paragraph{Low Realism Backgrounds: Generated backgrounds in Europe unrealistically portray cooking pots with outdoor backgrounds.}

Finally, we examine generated images whose \bgonly measurements never fall in the manifold of the real images.
This allows us to identify cases of poor generation realism. 
Using the \bgonly set-up, we inspect random examples of generated images created with the prompt \texttt{cooking pot in Europe} that do not fall in the real image manifold.
We observe that these generations often place the cooking pots in outdoor scenes, along brick walls or on a stony surface with foliage and dirt behind them, while real images of cooking pots depict them indoors.
Randomly selected examples are included in Figure \ref{fig:failure3}, with more in 
Appendix Figure \ref{fig:more-examples-failure-mode-3}.

\begin{figure}[ht]
    \centering
    \begin{minipage}[t]{0.5\textwidth}
        \centering
        \includegraphics[width=
    \textwidth]{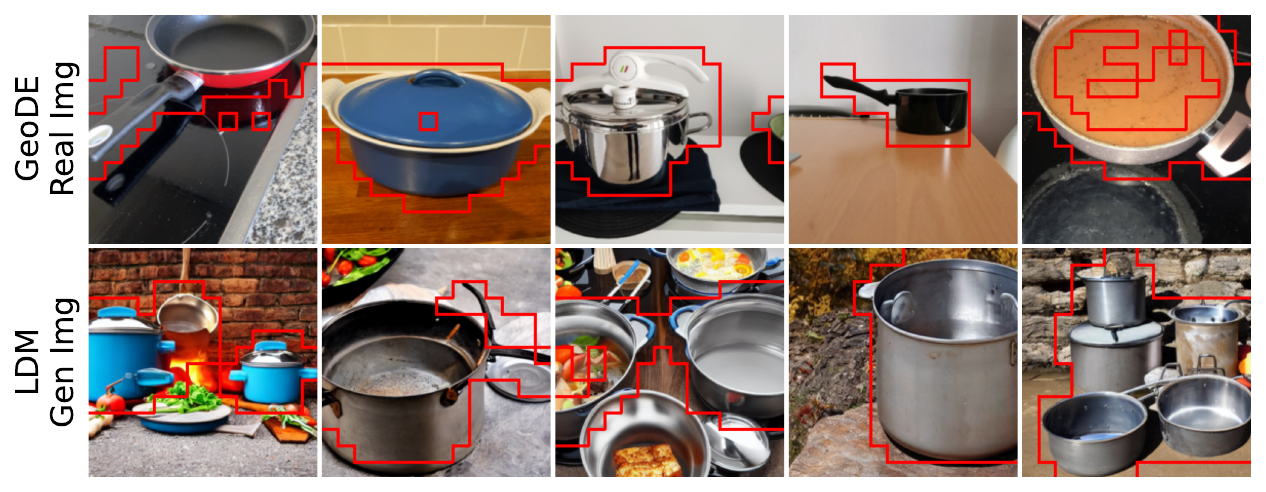}
    \end{minipage}
    \caption{\textbf{Generated images of cooking pots in Europe often unrealistically place the pots outdoors.}
    Depicted are generated images that, when measured with \bgonly setting, never fall in the manifold of real images.
    Red outlines show object/background decompositions.
    }
    \label{fig:failure3}
\end{figure}

\section{Early mitigations via new prompt template}

\begin{table}[ht]
\resizebox{\columnwidth}{!}{%
\begin{tabular}{c|cccc|cccc}
\multicolumn{1}{l|}{} & \multicolumn{4}{c|}{\textbf{\objonly}} & \multicolumn{4}{c}{\textbf{\bgonly}} \\
\multicolumn{1}{l|}{} &
  \multicolumn{1}{c}{\begin{tabular}[c]{@{}c@{}}Avg.\\ Prec.\end{tabular}} &
  \multicolumn{1}{c}{\begin{tabular}[c]{@{}c@{}}Worst \\ Prec.\end{tabular}} &
  \multicolumn{1}{c}{\begin{tabular}[c]{@{}c@{}}Avg.\\ Covg.\end{tabular}} &
  \multicolumn{1}{c|}{\begin{tabular}[c]{@{}c@{}}Worst \\ Covg.\end{tabular}} &
  \multicolumn{1}{c}{\begin{tabular}[c]{@{}c@{}}Avg.\\ Prec.\end{tabular}} &
  \multicolumn{1}{c}{\begin{tabular}[c]{@{}c@{}}Worst \\ Prec.\end{tabular}} &
  \multicolumn{1}{c}{\begin{tabular}[c]{@{}c@{}}Avg.\\ Covg.\end{tabular}} &
  \multicolumn{1}{c}{\begin{tabular}[c]{@{}c@{}}Worst \\ Covg.\end{tabular}} \\ \hline

\textbf{Orig.}       &   0.617    &  0.564     & 0.377      &  \textbf{0.352}     &  \textbf{0.481}     &  0.363     &  0.383    & 0.278     \\
\textbf{New}            &  \textbf{0.665}     &  \textbf{0.609}     & \textbf{0.390}      & 0.344      &  0.466     & \textbf{0.389}      &  \textbf{0.461}    & \textbf{0.423}     \\ \hline
\textbf{Delta}                 &  \textbf{0.048}     &  \textbf{0.045}           & \textbf{0.013}      &  -0.008     & -0.015      &  \textbf{0.026}    & \textbf{0.078} &     \textbf{0.145} \\
\textbf{Delta (\%)}                 &  \textbf{8\%}     &  \textbf{8\%}           & \textbf{3\%}      &  -2\%     & -3\%      &  \textbf{7\%}    & \textbf{20\%} &     \textbf{52\%}

\end{tabular}
}
\caption{\textbf{Prompting with region adjectives improves background diversity with little cost to object realism or diversity.}
Shown are \methodname values for the prompt templates \texttt{\{object\} in \{region\}} (\textbf{Orig.}) and \texttt{\{region adjective\} \{object\}} (\textbf{New}). Worst refers to the lowest performing region, primarily Africa and West Asia (see Figure \ref{fig:precision_coverage_regional_noun_objects}). 
}
\label{tab:mitigations_new}
\end{table}

Informed by our findings, we explore prompting as an early mitigation for regional disparities in generation. 
Because our analysis discovered backgrounds to be a driver of disparities, we study whether prompts that use adjective descriptors (``European bag'') rather than noun descriptors that could be construed as an instruction to depict the object \textit{in} the region (``bag in Europe'') reduce disparities in generations. 
We repeat experiments using the new prompting template.

Table \ref{tab:mitigations_new} shows results of the new prompting method in comparison with the original method, and Appendix Figure \ref{fig:precision_coverage_regional_noun_objects} shows a side-by-side comparison. 
We find that adjective-based prompting substantially improves background diversity (coverage), by 52\% for the worst region and 20\% on average. 
When we inspect random qualitative examples shown in Figure \ref{fig:main-disparities}, we see that the backgrounds tend to be more varied, with fewer stereotypical region representations not seen in the real dataset and more neutral and plain scenes. 
We see that this comes with a small improvement in worst-group background realism (precision) and a small drop of $3\%$ in average precision. 
In addition, the new prompting template leads to small improvements in realism and diversity in the depiction of objects on average.

\section{Conclusion}

In this work, we introduce the \methodname benchmark and use it to identify that a widely used LDM contributes to geographic disparities largely through the depiction of backgrounds. 
We identify specific failures like the lack of generation of red sedans for Africa and placing cooking pots outdoors for Europe. 
Finally, we experiment with an early attempt at mitigations via a new prompting strategy, demonstrating that \methodname can be useful in informing mitigations that can lead to up to a $52\%$ improvement in representation diveristy of backgrounds for the worst-performing region. 
We hope that this work paves the way for more fine-grained analysis and mitigation strategies addressing disparities in text-to-image generation.

\clearpage

\bibliography{example_paper}
\bibliographystyle{icml2024}

\newpage
\appendix
\onecolumn

\section{Social Impacts}
In this work, we introduced \methodname to better understand patterns of geographic disparities in text-to-image generative models.
The use of our metrics may have broader impacts in guiding the priorities of realism and representation diversity disparity mitigation efforts, especially along the axes of object and background representation. 
In addition, the nature of these metrics necessarily focuses analysis on certain geographic categorizations and groups.
While these criteria and groups have been shown to closely correspond to human prioritization when evaluating geographic representation \cite{hall2024geographic,basu2023inspecting}, we hope that our work does not preclude research into other methods for understanding geographic disparities and ensuring inclusivity in text-to-image geographic capabilities.

\section{Limitations}
While \methodname is useful in providing insights specific to object and background realism and representation  diversity in generated images, we discuss possible limitations of our method. 

First, it is possible that the feature extractor used for precision and coverage measurements is better for either backgrounds or objects, potentially biasing our results. 
However, we did not observe issues along these lines in our qualitative review of experimental results.

In addition, we can only apply \methodname for cases where the segmentation model and object detection pipeline works well. 
While we took steps to mitigate this risk by evaluating with the real images (for which we have ground truth information) and removing images and classes with significant failures, there may be unique failures for generated images. 
We analyze the small set of failure cases that we observed with the segmentation pipeline further in Appendix \ref{sec:app_methodology}.

Finally, automatic metrics are not sufficient in identifying the full breadth of human preference in image realism and representation diversity.
Thus, \methodname is best used accompanied by qualitative study.

\section{Additional Methodology Details}
\label{sec:app_methodology}

In this Section, we discuss additional details about our proposed methodology and related background material.

\subsection{Precision and coverage definitions}
\subsubsection{Precision or Realism}
Precision measures how close the generated images are to the real images from the reference dataset. 
We calculate the precision by following the approach in \citet{kynkäänniemi2019improved} and \citet{hall2023dig}, where we determine the proportion of the generated images that lie within the manifold, i.e.\ $k$-th nearest neighbor distance of at least one of the real images in a pre-defined feature space. 
The existing work \cite{hall2023dig} calculates this $k$-th nearest neighbor distance in the InceptionV3 \citep{DBLP:journals/corr/SzegedyVISW15} feature space. 
The formula for calculating the precision is given by previous works \cite{hall2023dig,kynkäänniemi2019improved} as:
\begin{equation}
    \textrm{P}(\mathcal{D}_r, \mathcal{D}_g) = \frac{1}{|\mathcal{D}_g|} \sum_{i=1}^{|\mathcal{D}_g|} \mathbbm{1}_{h_g^{(i)} \in  \textrm{manifold}(\mathcal{D}_r)},
\end{equation}
\textrm{P}
where $\mathcal{D}_r = \{h_r^{(j)}\}$ is dataset of reference images features, $\mathcal{D}_g = \{h_g^{(i)}\}$ is dataset of generated images features.

\subsubsection{Coverage or Diversity}
Diversity is a measure of how representative the generated images dataset is of the reference dataset. 
In other words, it is the proportion of the reference images that have at least one of the generated images within their manifold in a pre-defined feature space.
We use this definition of coverage proposed in \citet{naeem2020reliable}, to measure the diversity of generated images. We compute the coverage following the prior works \citep{hall2023dig,naeem2020reliable} as follows: 

\begin{equation}
    \textrm{C}(\mathcal{D}_r, \mathcal{D}_g) = \frac{1}{|\mathcal{D}_r|} \sum_{j=1}^{|\mathcal{D}_r|} \mathbbm{1}_{\exists \:  i \:  \mathrm{s.t.}\: h_g^{(i)} \in \mathrm{manifold}\left(h_r^{(j)}\right)}.
\end{equation}

\subsection{Language-based object and background segmentation}

To build our \methodname metrics, we first perform an automatic object-background segmentation using LangSAM,\footnote{https://github.com/luca-medeiros/lang-segment-anything} a state-of-the-art segmentation tool. LangSAM automatically segments images in two stages. The first stage uses GroundingDino  \cite{liu2023grounding} to generate a bounding box for an image given the object prompt from the text prompt we provide (e.g. ``bag''). The second stage uses the bounding box to prompt a segmentation model SAM \citep{kirillov-etal-2023-segment}.  

\begin{figure*}
  \centering
    \includegraphics[width=0.5\textwidth]{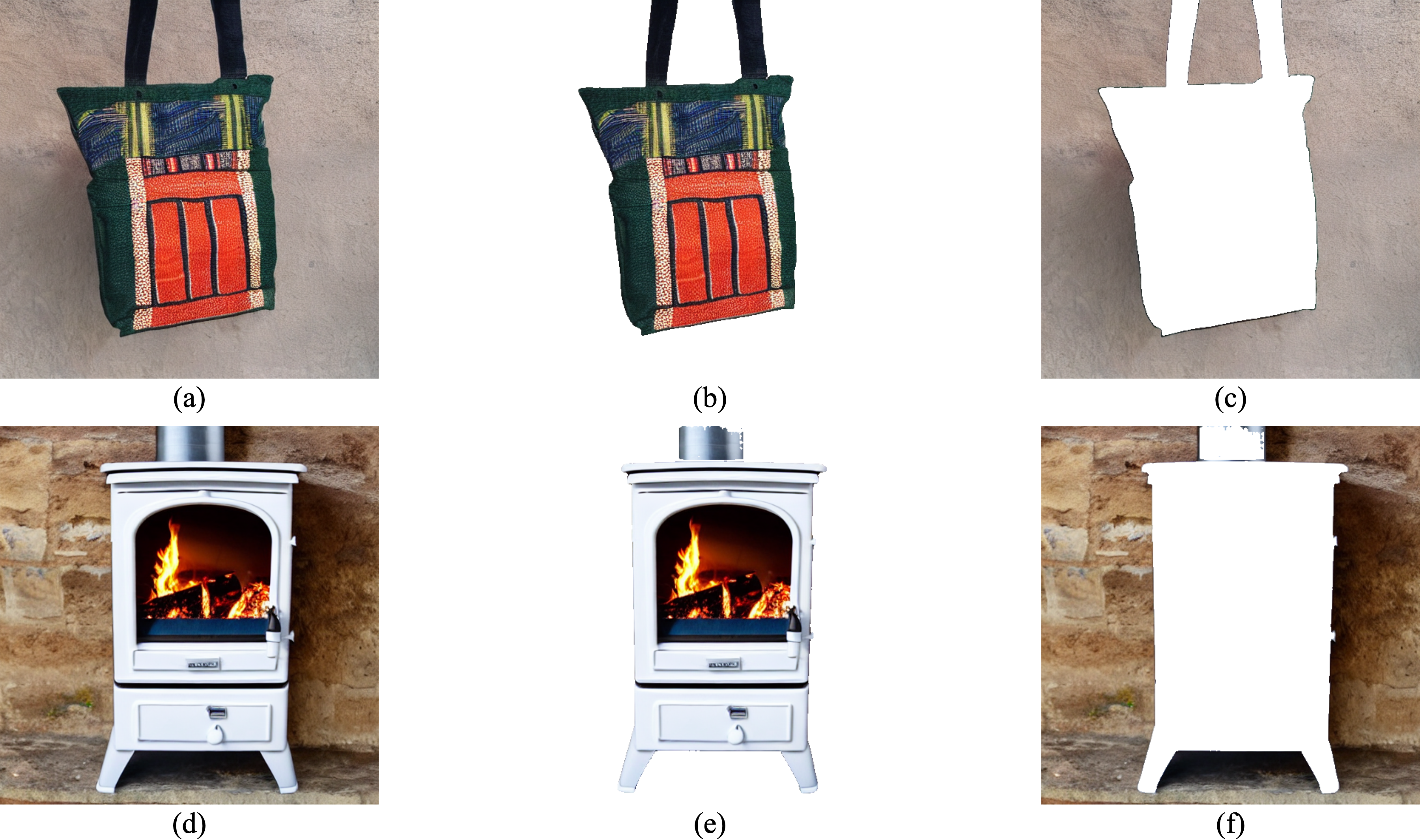}
  \caption{Illustration of the image segmentation using LangSAM: full images (a, d), segmented objects (b, e) and segmented backgrounds (c, f).}
   \label{fig:Lang_SAM}  
\end{figure*}

Once we have the segmentation masks, we fill in the removed components of the image. For object-segmented images, the background is whitened. For background-segmented images, the object is whitened. Figure \ref{fig:Lang_SAM} illustrates the resulting object and background segmented images using LangSAM.

\subsubsection{Data Filtering}

The real image dataset, GeoDE \cite{ramaswamy2022geode} has $27$ classes of objects with at least $170$ images in each of the 6 regions. We found  $3$ classes in GeoDE had over 100 segmentation failures out of 1020 images ($170\times6$ regions), and we excluded these classes from analyses. The excluded classes are: ``cleaning equipment'', ``hand soap'' and ``light fixture''. Hence, we used only the resulting $24$ classes for our analysis, namely:
tree,
hat,
cooking pot,
jug,
toothbrush,
bag,
waste container,
bicycle,
chair,
stove,
dog,
car,
plate of food,
spices,
medicine,
hairbrush comb,
toy,
light switch,
front door,
storefront,
lighter,
dustbin,
toothpaste toothpowder, and
candle. 

\subsection{Feature extractors}
Initially, we experimented with convolutional neural network (CNN) based models like Inception-v3 \cite{DBLP:journals/corr/SzegedyVISW15}.
However, a significant drawback of CNNs is that they consider the entire image for feature extraction. In segmented images, large portions of the image are replaced with a specified pixel value, affecting the resulting features that are extracted. With patch-based models, we avoid this problem by only selecting patches within the segmentation.

\section{Additional findings}
\label{app:additional_findings}

In this Section, we include additional experimental results.

\subsection{Effects of new prompting strategy}

\begin{figure}
    \centering
    \begin{minipage}[t]{0.8\textwidth}
        \centering
        \includegraphics[width=
    \textwidth]{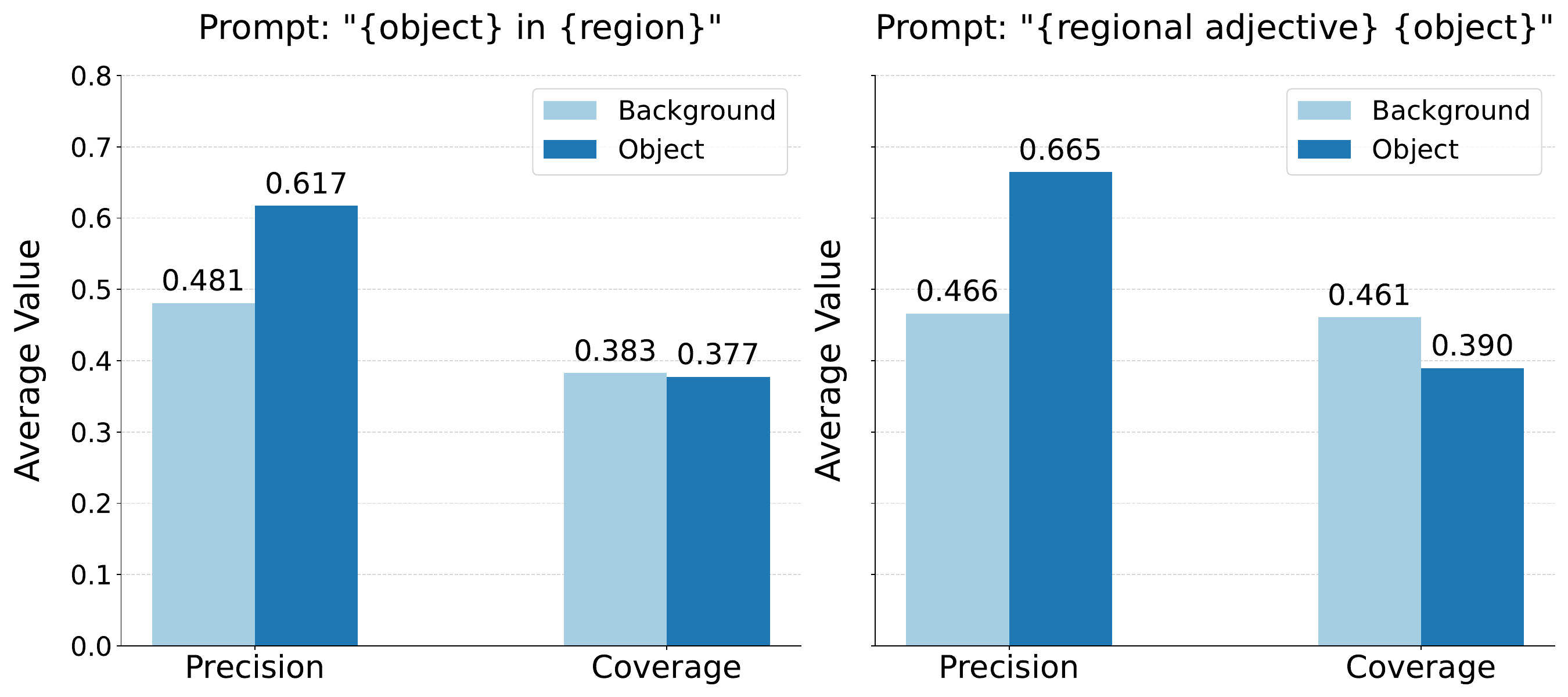}
    \end{minipage}
    \caption{Precision and coverage measurements averaged over all regions for \objonly and \bgonly set-ups for \texttt{\{object\} in \{region\}} generation prompt (left) and \texttt{\{regional adjective\} \{object\}} prompt (right).
    }
    \label{fig:pc_avgd_combined}
\end{figure}

\begin{figure*}
    \centering
    \includegraphics[width=0.7\textwidth]{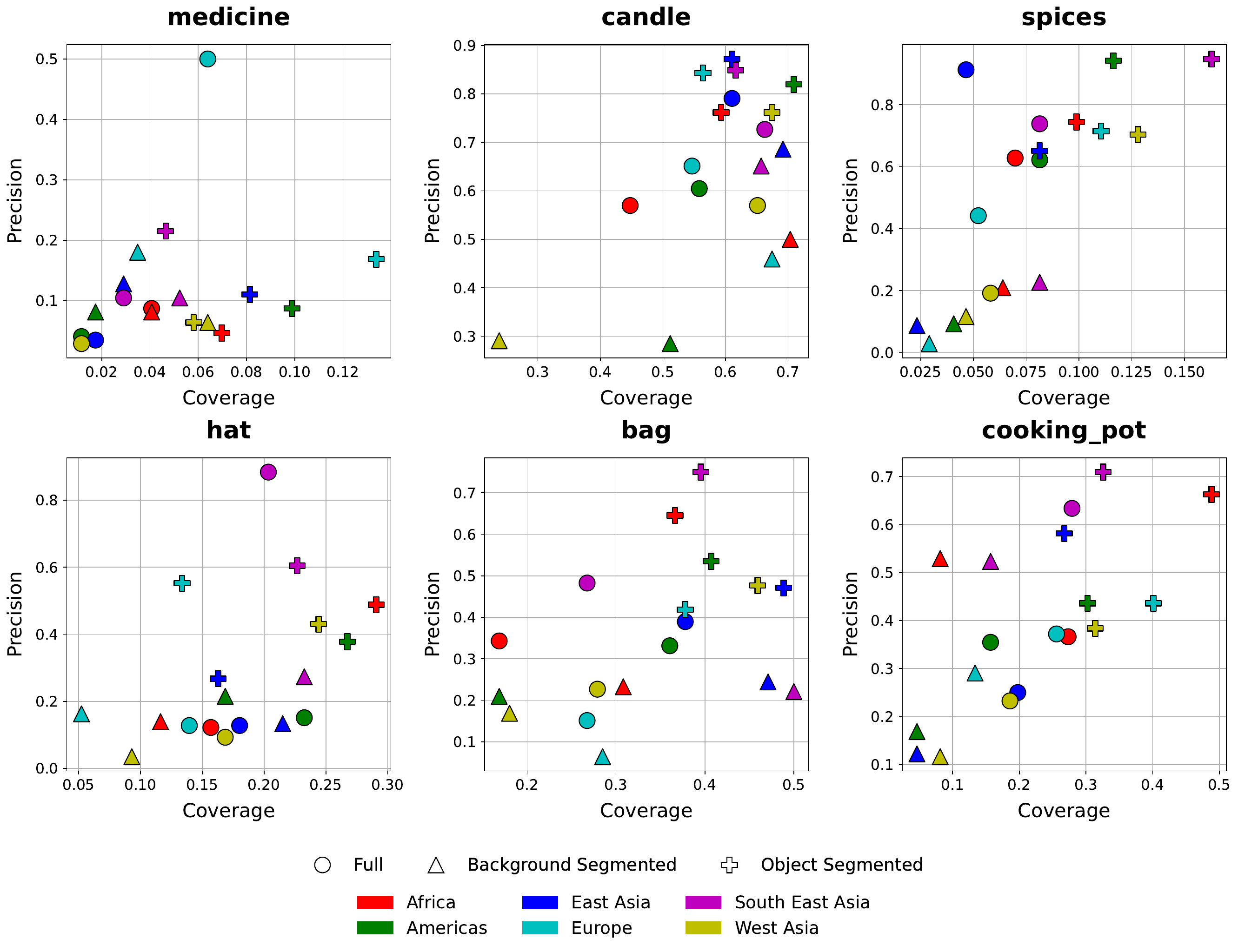}
    \caption{Object-wise \methodname with prompt template \texttt{\{object\} in \{region\}}. Note axes scale differences between plots.}
\label{fig:precision_coverage_Object_regional_noun_objects}
\end{figure*}

\begin{figure*}
    \centering
    \includegraphics[width=0.7\textwidth]{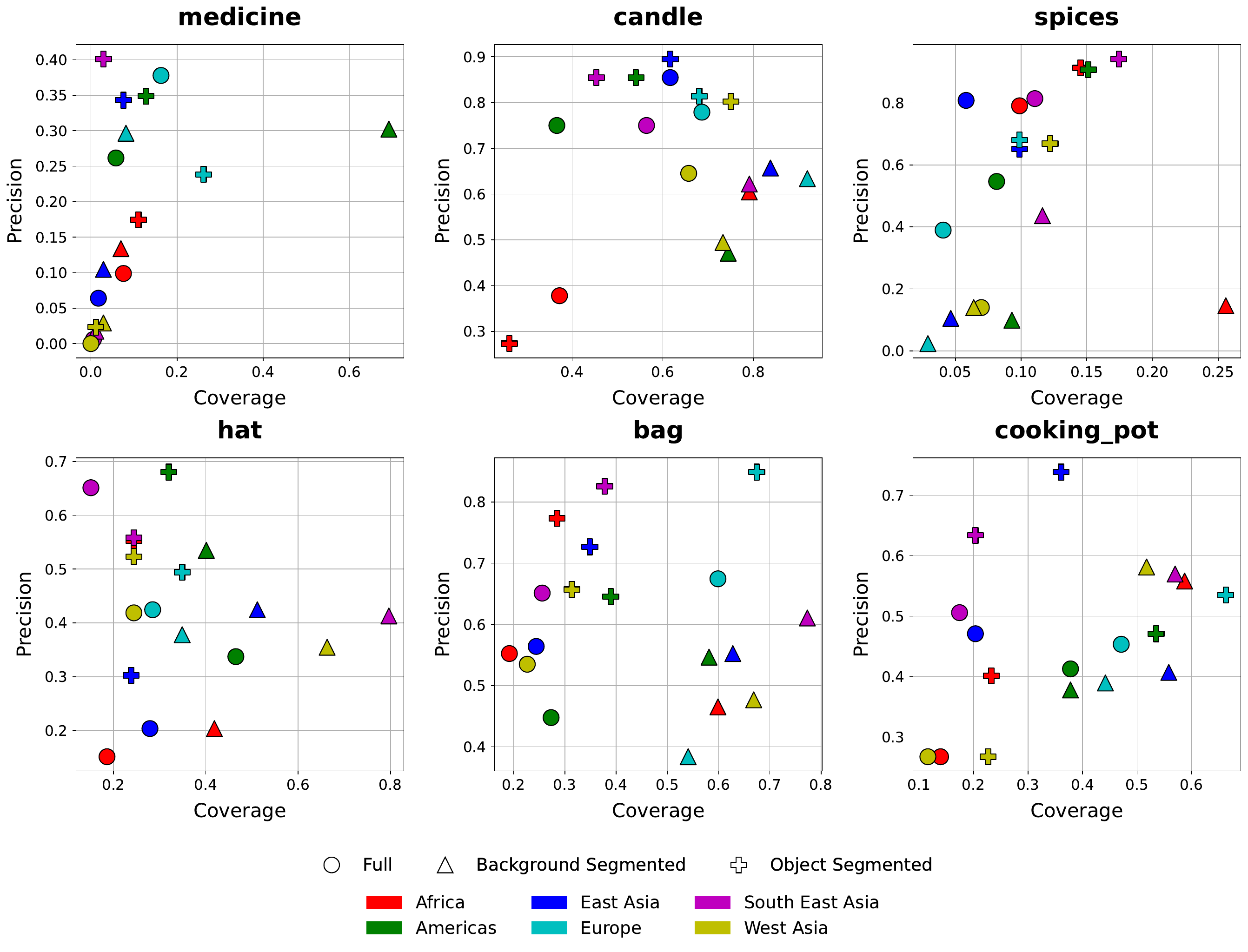}
    \caption{Object-wise \methodname with prompt template \texttt{\{regional adjective\} \{object\}}. Note axes scale differences between plots.}
    \label{fig:precision_coverage_regional_adjective_objects}
\end{figure*}

Figure \ref{fig:pc_avgd_combined} shows results averaged over all regions comparing old prompting structure -- \texttt{\{object\} in \{region\}}, and the new prompting structure -- \texttt{\{regional adjective\} \{object\}}.
We also include object-specific measurements with the old and new prompting strategies in Figures \ref{fig:precision_coverage_Object_regional_noun_objects} and \ref{fig:precision_coverage_regional_adjective_objects} respectively.

\subsection{Full image results}

\begin{figure*}
    \centering
    \includegraphics[width=0.8\textwidth]{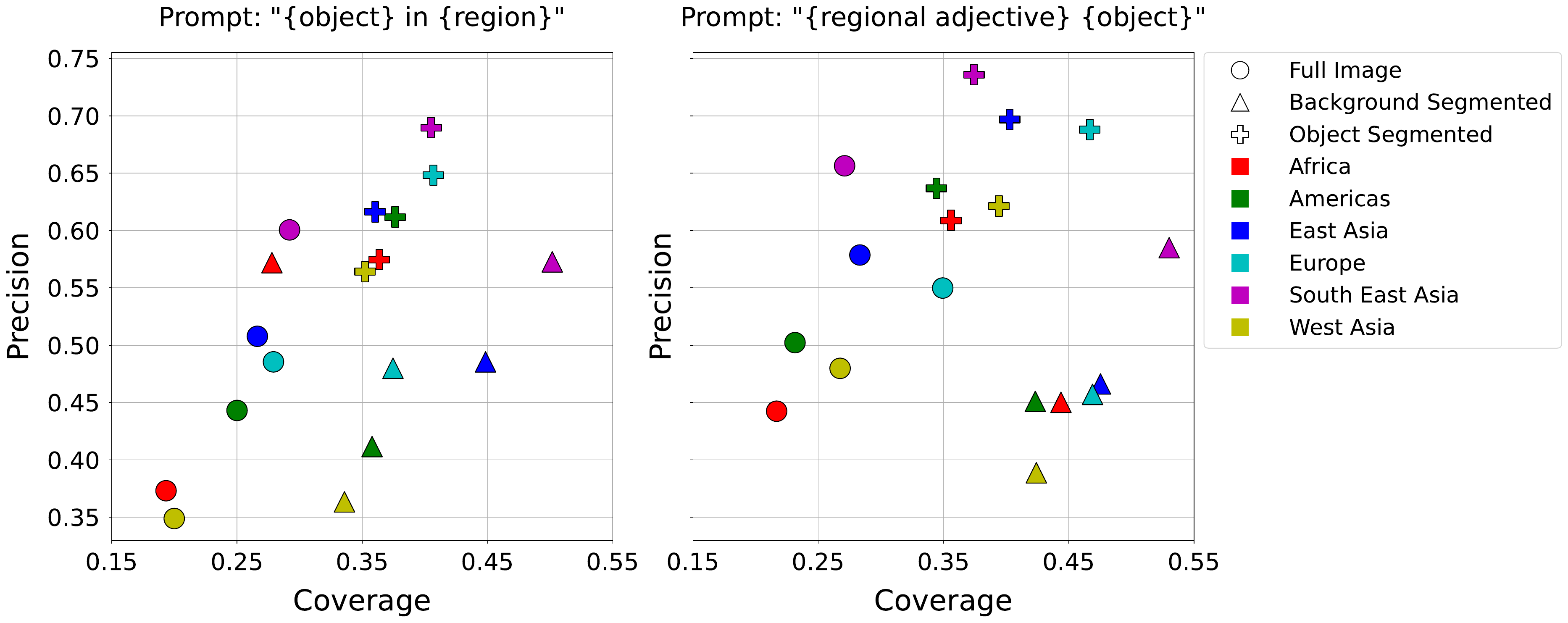}
    \caption{\methodname metrics comparing the two prompting strategies.}
    \label{fig:precision_coverage_regional_noun_objects}
\end{figure*}

We show a side-by-side comparison of both prompting strategies for \fullimg, \objonly, and \bgonly measurements in Figure \ref{fig:precision_coverage_regional_noun_objects}.

\subsection{Analyzing the precision \bgonly score for Africa}
\begin{figure}
    \centering
    \begin{minipage}[t]{0.49\textwidth}
        \centering
        \includegraphics[width=
    \textwidth]{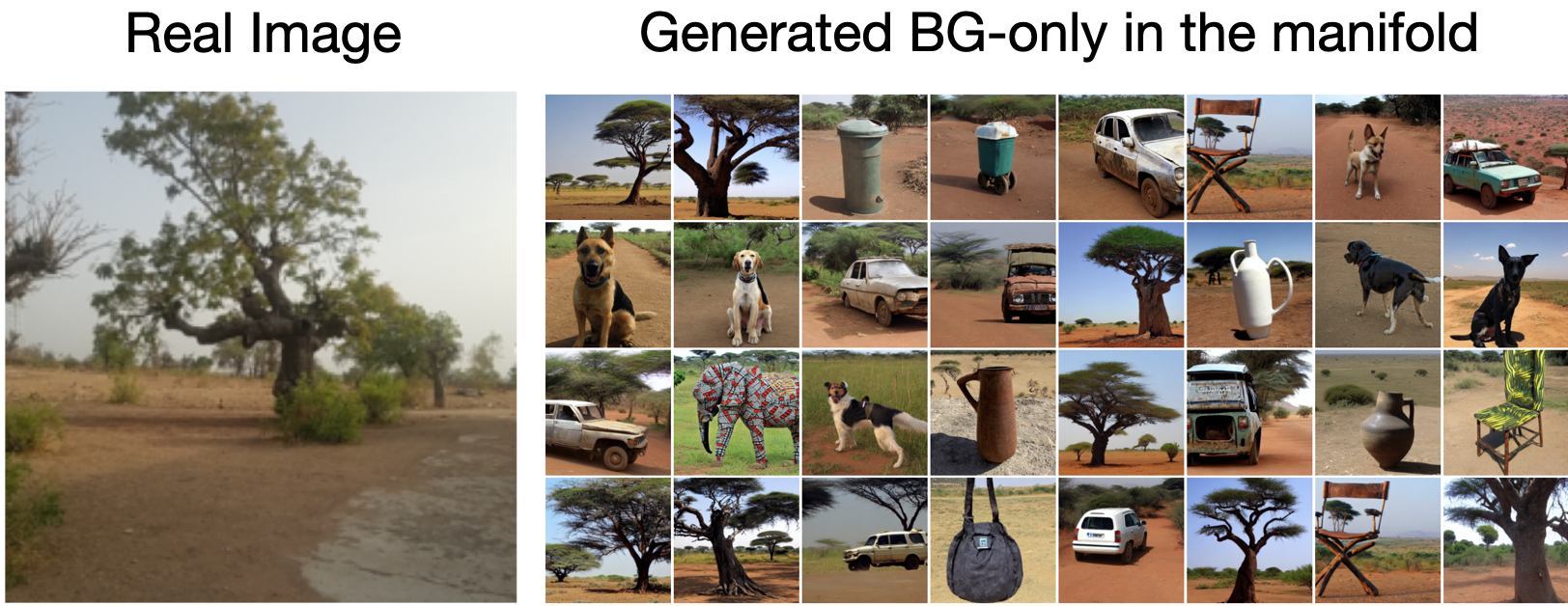}
    \end{minipage}
    \hfill\vline\hfill
    \begin{minipage}[t]{0.49\textwidth}
        \centering
        \includegraphics[width=
    \textwidth]{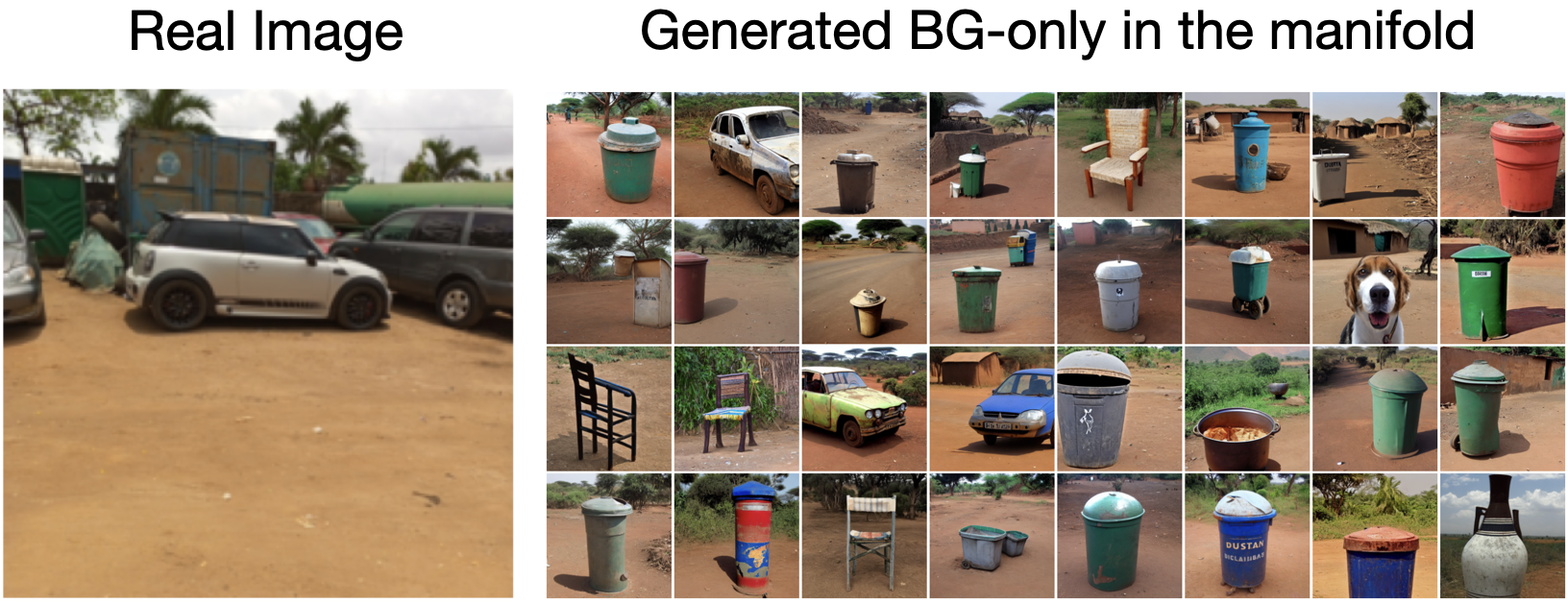}
    \end{minipage}
    \caption{Examples of the reference image from African region which have many generated images falling into the real image manifold in \bgonly setting.}
    \label{fig:more-fake-imgs-in-real-africa}
\end{figure}

From Figure \ref{fig:plot_pc_combined_all_regions_section_4_2}, we observe that Africa has high precision in \bgonly setting.
We conjecture that this is because most of the generated image backgrounds are stereotypical and resemble sandy and plain desert areas,
while most of the images in the reference dataset (GeoDE) have diverse set of backgrounds such as paved roads, kitchen floors, and grass areas. 
There are very few images of Africa in the reference dataset which contain stereotypical backgrounds and, as a result, most of the generated images fall into the manifolds of \bgonly setting of these few images.
As an illustration, the real images in Figure \ref{fig:more-fake-imgs-in-real-africa} have more than $50$ generated images included in its manifold in \bgonly setting. 
However, other more realistic backgrounds do not have any associated generated image in their manifold. 
Hence, we hypothesize that a minority of real images led to high precision and low coverage scores for \bgonly measurement in Africa.

\subsection{Additional visual examples}

\begin{figure*}
    \centering
    \includegraphics[width=\textwidth]{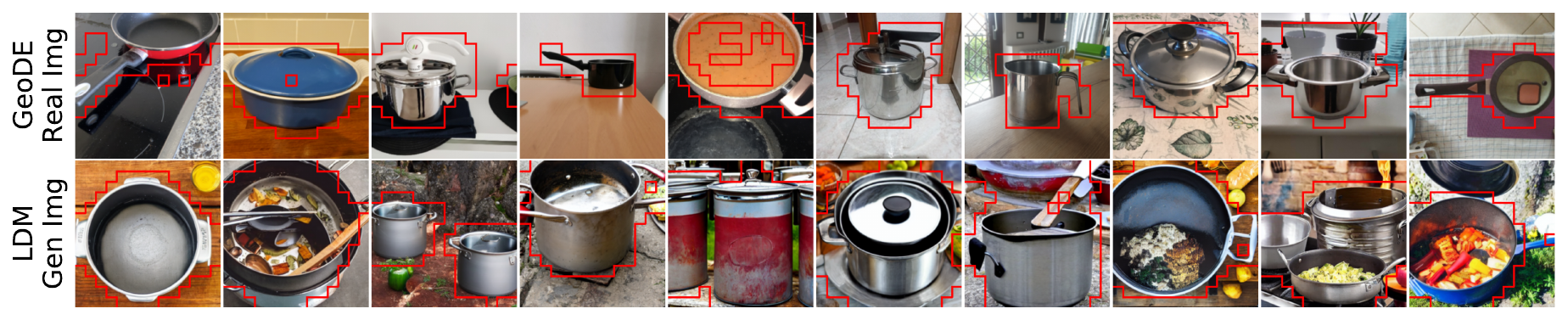}
    \caption{Additional examples of the generated images for \texttt{cooking pot in Europe} template, illustrating the low realism for background generation.}
    \label{fig:more-examples-failure-mode-3}
\end{figure*}

Figure \ref{fig:more-examples-failure-mode-3} contains additional examples of generated images unrealistically placing cooking pots outdoors.

\end{document}